\documentclass[10pt,onecolumn,twoside]{article}

\usepackage{fullpage}

\usepackage{amsmath}
\usepackage{amssymb}
\usepackage{graphicx}
\usepackage{amsthm}
\usepackage{cite}
\usepackage{float}
\usepackage{bm}
\usepackage{url}
\usepackage{algorithmic}
\usepackage{algorithm}
\usepackage{authblk} % affiliations

\def\defn{\,\triangleq\,}
\def\half{{\textstyle\frac{1}{2}}}
\def\argmin{\mathop{\mathrm{arg\,min}}} % Argument of a minimization
\def\prox{\mathrm{prox}}
\def\real{\mathrm{Re}}

\def\uin{u_{\text{\tiny in}}}
\def\uscat{u_{\text{\tiny sc}}}

\def\argmin{\mathop{\mathrm{arg\,min}}}

\def\C{\mathbb{C}}
\def\R{\mathbb{R}}

\def\ybf{{\mathbf{y}}}
\def\zbf{{\mathbf{z}}}
\def\gbf{{\mathbf{g}}}

\def\fbf{{\mathbf{f}}}
\def\ubf{{\mathbf{u}}}
\def\bbf{{\mathbf{b}}}
\def\rbf{{\mathbf{r}}}
\def\vbf{{\mathbf{v}}}

\def\xbm{{\bm{x}}}

\def\d{{\, \mathrm{d}}}

\def\im{{\mathrm{j}}}

\def\Hrm{{\mathrm{H}}}

\def\fbfhat{{\widehat{\mathbf{f}}}}

\def\ubfbar{{\bar{\ubf}}}

\def\Gbf{{\mathbf{G}}}
\def\Hbf{{\mathbf{H}}}
\def\Dbf{{\mathbf{D}}}

\def\Rcal{\mathcal{R}}
\def\Ocal{\mathcal{O}}
\def\Dcal{\mathcal{D}}

\def\Fcal{\mathcal{F}}

\theoremstyle{definition}

\begin{document}

%%%%%%%%%%%%%%%%%%%%%%%%%%%%%%%%%%%%%%%%%%%%%
%% Title
%%%%%%%%%%%%%%%%%%%%%%%%%%%%%%%%%%%%%%%%%%%%%

\title{A Recursive Born Approach to Nonlinear Inverse Scattering}

%%%%%%%%%%%%%%%%%%%%%%%%%%%%%%%%%%%%%%%%%%%%%
%% Authors
%%%%%%%%%%%%%%%%%%%%%%%%%%%%%%%%%%%%%%%%%%%%%

\author{Ulugbek~S.~Kamilov%
\thanks{U.~S.~Kamilov (email: kamilov@merl.com),  D.~Liu (email: liudh@merl.com), H.~Mansour (email: mansour@merl.com), and P.~T.~Boufounos (email: petrosb@merl.com) are with Mitsubishi Electric Research Laboratories (MERL), 201 Broadway, Cambridge,
MA 02140, USA.}
\hspace{0.05em}, Dehong~Liu, Hassan~Mansour, and Petros~T.~Boufounos}

\maketitle %% required

%%%%%%%%%%%%%%%%%%%%%%%%%%%%%%%%%%%%%%%%%%%%%
%% Abstract
%%%%%%%%%%%%%%%%%%%%%%%%%%%%%%%%%%%%%%%%%%%%%

\begin{abstract}
The Iterative Born Approximation (IBA) is a well-known method for
describing waves scattered by semi-transparent objects. In this
paper, we present a novel nonlinear inverse scattering method that
combines IBA with an edge-preserving total variation (TV)
regularizer. The proposed method is obtained by relating iterations of
IBA to layers of a feedforward neural network and developing a
corresponding error backpropagation algorithm for efficiently
estimating the permittivity of the object. Simulations illustrate
that, by accounting for multiple scattering, the method successfully
recovers the permittivity distribution where the traditional
linear inverse scattering fails.
\end{abstract}

%%%%%%%%%%%%%%%%%%%%%%%%%%%%%%%%%%%%%%%%%%%%%
%% Section 1
%%%%%%%%%%%%%%%%%%%%%%%%%%%%%%%%%%%%%%%%%%%%%

\section{Introduction}
\label{Sec:Introduction}

Knowledge of the spatial distribution of the permittivity within
an object is important for many applications since it enables the
visualization of the internal structure and physical properties of the
object. Measurements of the permittivity are typically obtained by
first illuminating the object with a known incident wave and recording
the resulting scattered waves with sensors located outside the
object. The spatial map of permittivity is reconstructed from the
measurements, using computational inverse scattering methods that rely
on physical models describing the object-wave interaction.

Traditional approaches to inverse scattering formulate the task as a
linear inverse problem by establishing a linear relationship between
the permittivity and the scattered wave. The linear model can be
obtained by assuming a straight-ray propagation of
waves~\cite{Kak.Slaney1988}, or by adopting more refined
single-scattering models based on the first Born~\cite{Wolf1969} or
Rytov approximations~\cite{Devaney1981}. Once linearized, the problem
can be efficiently solved using the preferred regularized
reconstruction methods, typically based on sparsity and iterative
optimization~\cite{Bronstein.etal2002, Sung.Dasari2011, Kim.etal2014}.

Recent experimental results indicate that the resolution and quality
of the reconstructed permittivity is improved when nonlinear
physical models are used instead of traditional linear
ones~\cite{Simonetti2006,Simonetti.etal2008,Maire.etal2009,Kamilov.etal2015}. In
particular, nonlinear physical models can account for multiple
scattering and provide a more accurate interpretation of the measured
data at the cost of increased computational complexity of the
reconstruction.

In this paper, we develop a new computational imaging method to
reconstruct the permittivity distribution of an object from
transmitted or reflected waves. Our method is based on a nonlinear
physical model that can account for multiple scattering in a
computationally efficient way. Specifically, we propose to interpret
the iterations of the iterative Born approximation (IBA) as layers of
a feedforward neural network. This formulation leads to an efficient
error backpropagation algorithm used to evaluate the gradient of the
scattered field with respect to the permittivity, thus enabling
the recovery of the latter from a set of measured scattered
fields. The quality of the final estimate is further enhanced by
regularizing the solution with an edge-preserving total variation (TV)
penalty. Our simulations indicate that the proposed method accurately
models scattering without prohibitive computational overhead, and
successfully recovers the object where the traditional linear
approaches fail.

%%%%%%%%%%%%%%%%%%%%%%%%%%%%%%%%%%%%%%%%%%%%%
%% Section 2
%%%%%%%%%%%%%%%%%%%%%%%%%%%%%%%%%%%%%%%%%%%%%

\section{Related Work}
\label{Sec:RelatedWork}

A comprehensive review of nonlinear inverse scattering, with detailed description of standard algorithms, is available in the book-chapter by van den Berg~\cite{vandenBerg2002}. 

Three common approaches to nonlinear inverse scattering are iterative Born~\cite{Tijhuis1989,Wang.Chew1989,Chew.Wang1990}, modified gradient~\cite{Kleinman.vandenBerg1992,Kleinman.vandenBerg1993,Belkebir.Sentenac2003}, and contrast-source inversion methods~\cite{vandenBerg.Kleinman1997}. All these methods attempt to iteratively minimize the discrepancy between the actual and predicted measurements, while enforcing the consistency of the fields inside the object. The actual optimization is performed in an alternating fashion by first updating the permittivity for a fixed field, and then updating the field for a fixed permittivity. The difference between the methods is in the actual computation of the updates.

Recently, the beam propagation method (BPM) was proposed for
performing nonlinear inverse scattering in
transmission~\cite{Tian.Waller2015,Kamilov.etal2015,Waller.Tian2015,Kamilov.etal2016}. BPM-based
methods circumvent the need to solve an explicit optimization problem
for the internal field, by instead numerically propagating the field
slice-by-slice through the object. It was shown that BPM can be
related to a neural network where each layer corresponds to a spatial
slice of the object. An efficient backpropagation algorithm can thus
be derived to reconstruct the object from the
measurements~\cite{Kamilov.etal2016}.

This paper substantially extends the BPM-based method by relying on
IBA to formulate the nonlinear physical model. One main advantage of
the proposed formulation over BPM is that it allows reconstruction
when measuring reflections. Additionally, unlike alternating
minimization schemes that assume a linear problem for a given internal
field, our method directly optimizes the nonlinear model in a
tractable way using error backpropagation.

%%%%%%%%%%%%%%%%%%%%%%%%%%%%%%%%%%%%%%%%%%%%%
%% Section 3
%%%%%%%%%%%%%%%%%%%%%%%%%%%%%%%%%%%%%%%%%%%%%

\section{Main Results}
\label{Sec:MainResults}

The method presented here can be generalized to a majority of
tomographic experiments in transmission or reflection. For simplicity
of derivations, we ignore absorption by assuming a real permittivity; 
however, the equations can be generalized to handle complex
permittivities. We additionally assume coherent measurements,
i.e., that both the amplitude and phase of the scattered wave are
recorded at the sensor locations.

%%%%%%%%%%%%%%%%%%%%%%%%%%%%%%%%%%%%%%%%%%%%%
%% scene
%%%%%%%%%%%%%%%%%%%%%%%%%%%%%%%%%%%%%%%%%%%%%

\begin{figure}[t]
\centering\includegraphics[width=6.5cm]{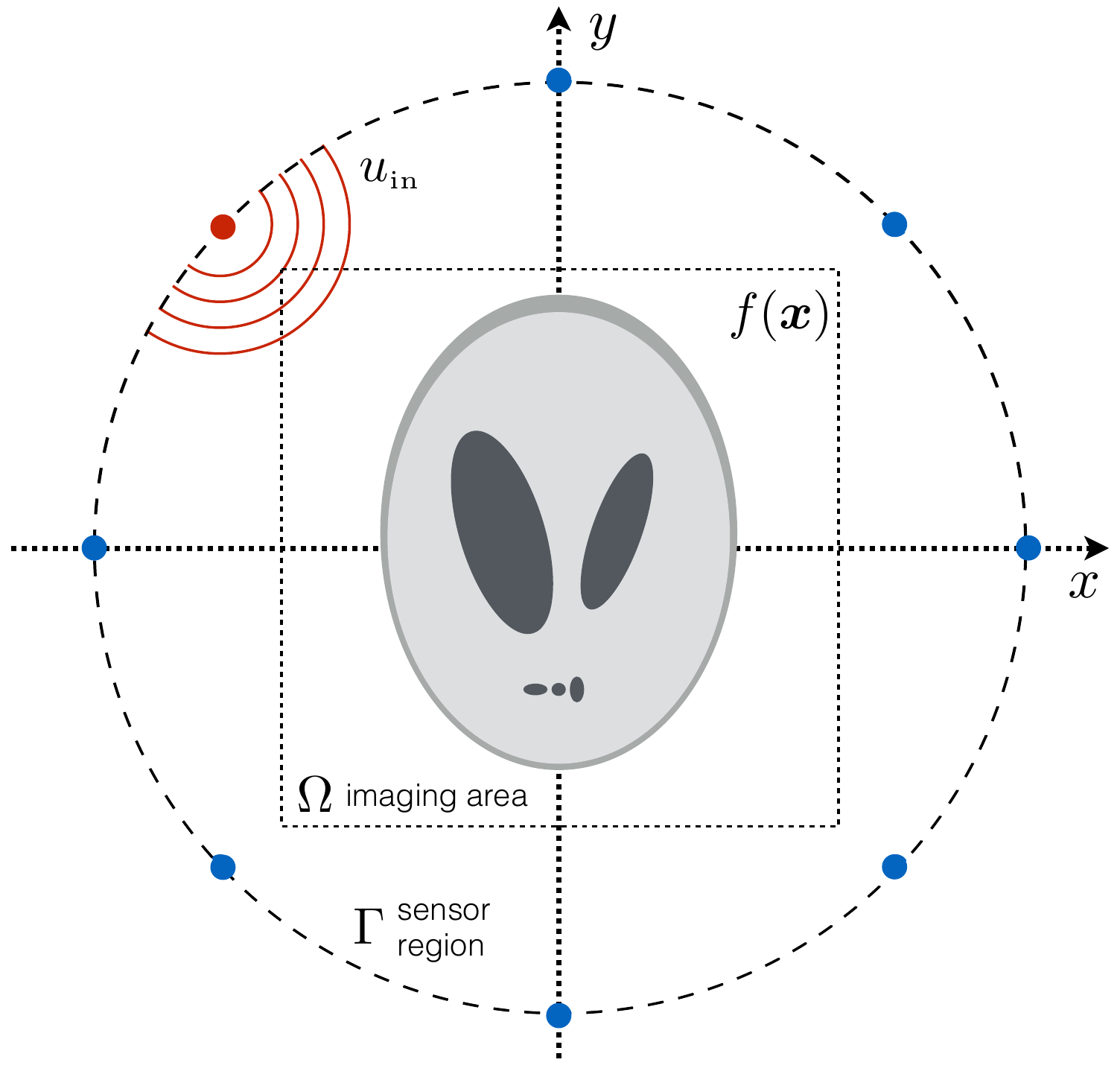}
\caption{Schematic representation of a scattering scenario. An object
  with a scattering potential $f(\xbm)$, $\xbm \in \Omega$, is
  illuminated with an input wave $\uin$, which interacts with the
  object and results in the scattered wave $\uscat$ at the sensor
  region $\Gamma$. The scattered wave is measured and used to
  computationally form an image of $f$.}
\label{Fig:ScatteringScenario}
\end{figure}

%%%%%%%%%%%%%%%%%%%%%%%%%%%%%%%%%%%%%%%%%%%%%
%% Formulation of the problem
%%%%%%%%%%%%%%%%%%%%%%%%%%%%%%%%%%%%%%%%%%%%%

\subsection{Problem Formulation}
\label{Sec:Formulation}

Consider the two-dimensional (2D) scattering problem illustrated in
Fig.~\ref{Fig:ScatteringScenario}, where an object of real permittivity 
distribution $\epsilon(\xbm)$, with $\xbm = (x, y) \in \Omega$, is
immersed into the background medium of permittivity $\epsilon_b$. The line sources
that generate the electromagnetic excitation and the sensors
collecting the data are located in the sensor region $\Gamma \subseteq
\R^2$. By assuming, and ignoring, a time dependence $\exp(\im \omega
t)$, the incident electric field created by the $\ell$th source,
located at $\xbm_\ell \in \Gamma$, is given by
\begin{equation}
\uin(\xbm) = A\,\frac{\im}{4} H_0^{(2)}(k_b \|\xbm-\xbm_\ell\|_{\ell_2}),
\end{equation}
for all $\xbm \in \R^2$, where $A$ is the strength of the source,
$H_0^{(2)}$ is the zero-order Hankel function of the second kind, $k_b
= k_0 \sqrt{\epsilon_b}$ is the wavenumber in the background medium, ${k_0 =
  \omega/c_0 = 2\pi/\lambda}$ is the wavenumber in free space,
$\lambda$ is the wavelength, and $c_0 \approx 3 \times 10^8$ m/s. In
the subsequent derivations, we consider the scenario of a single
illumination and drop the indices $\ell$. The generalization to an
arbitrary number of illuminations $L$ is straightforward.

The Lippmann-Schwinger integral equation describes the relationship
between the permittivity and the wave-field~\cite{vandenBerg2002}
\begin{equation}
\label{Eq:LippmanSchwinger}
u(\xbm) = \uin(\xbm) + \int_{\Omega} g(\xbm-\xbm^\prime) f(\xbm^\prime) u(\xbm^\prime) \d \xbm^\prime,
\end{equation}
for all $\xbm \in \Omega$, where we define the \emph{scattering potential}
\begin{equation}
f(\xbm) \defn k_b^2(\epsilon_b-\epsilon(\xbm))
\end{equation}
and the Green's function for the homogeneous medium
\begin{equation}
g(\xbm) \defn \frac{\im}{4} H_0^{(2)}(k_b \|\xbm\|_{\ell_2}).
\end{equation} 
Similarly, the scattered field in the sensor region can be expressed as
\begin{equation}
\label{Eq:ScatteredField}
\uscat(\xbm) = u(\xbm)-\uin(\xbm) = \int_{\Omega} g(\xbm - \xbm^\prime) f(\xbm^\prime) u(\xbm^\prime) \d \xbm^\prime
\end{equation}
for any $\xbm \in \Gamma$.
Note that the integrals~\eqref{Eq:LippmanSchwinger} and~\eqref{Eq:ScatteredField} extend only over $\Omega$ because the scattering potential $f$ is zero for all $\xbm \notin \Omega$.

The goal of inverse scattering is to estimate $f$, which is equivalent to $\epsilon$, given $M$ measurements of $\{\uscat(\xbm_m)\}_{m \in [1\dots M]}$ in the sensor region $\Gamma$. 
%In practice, one considers $L$ illuminations of the object and records $M$ spatial samples for each illumination.

%%%%%%%%%%%%%%%%%%%%%%%%%%%%%%%%%%%%%%%%%%%%%
%% Iterative Born approximation
%%%%%%%%%%%%%%%%%%%%%%%%%%%%%%%%%%%%%%%%%%%%%

\subsection{Iterative Born Approximation}
\label{Sec:IterativeBorn}

At first glance, it might seem that~\eqref{Eq:ScatteredField} directly
provides a linear relationship between $f$ and $\uscat$, which can be
used to solve the problem. However, the nonlinear nature of the
inverse scattering becomes evident if one realizes that the internal
field ${u = \uin + \uscat}$ in~\eqref{Eq:ScatteredField} depends on
$\uscat$.

We now consider a $K$-term iterative Born approximation
(IBA)~\cite{Born.Wolf2003} of the total
field~\eqref{Eq:LippmanSchwinger}
\begin{equation}
\label{Eq:IterativeBorn}
u_k(\xbm) = \uin(\xbm) + \int_\Omega g(\xbm-\xbm^\prime) f(\xbm^\prime) u_{k-1}(\xbm^\prime) \d \xbm,
\end{equation}
where $\xbm \in \Omega$, $u_0 = 0$, and $k = 1, 2, \dots, K$. When $K = 1$, eq.~\eqref{Eq:IterativeBorn} reduces to the well-known first-Born approximation, which assumes a single scattering from $f$ by approximating $u(\xbm)$ with the incident field $\uin(\xbm)$. For $K = 2$, the approximation of the total field is improved by taking into account the second scattering due to an additional interaction between the object and the field. For higher values of $K$ the approximation is further improved by accounting for multiple scattering of order $K$ (see also the discussion on multiple scattering by Born and Wolf~\cite{Born.Wolf2003}).

In the context of general theory of integral equations, the iterations of type~\eqref{Eq:IterativeBorn} are known as Liouville-Neumann series~\cite{Arfken.Weber2005a}. The sufficient condition for convergence states that the norm of the integral operator acting on the field should be less than unity. While this implies that IBA might diverge for large permittivity contrasts, it is still expected to work on a wide range of contrasts where the linearized models fail, which is corroborated by our simulations in Section~\ref{Sec:Experiments}.

%%%%%%%%%%%%%%%%%%%%%%%%%%%%%%%%%%%%%%%%%%%%%
%% Neural network
%%%%%%%%%%%%%%%%%%%%%%%%%%%%%%%%%%%%%%%%%%%%%

\begin{figure}[t]
\centering\includegraphics[width=8.5cm]{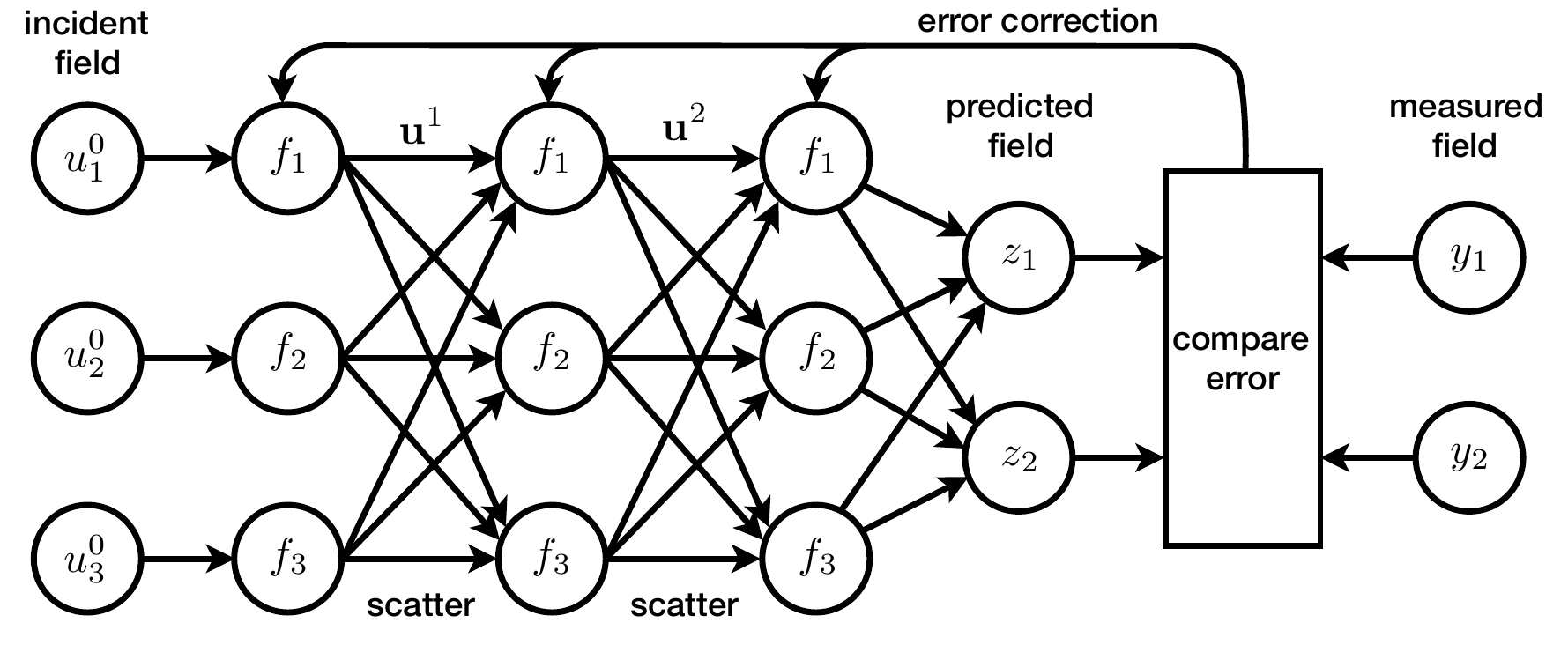}
\caption{Schematic representation of the multiple scattering as a neural network for $K = 2$, $N = 3$, and $M = 2$. The scattering potential $\fbf \in \R^N$ plays the role of nonlinearities, while matrices $\Hbf \in \C^{M \times N}$ and $\Gbf \in \C^{N \times N}$ represent weights of the network. The error backpropagation algorithm described here allows to efficiently estimate $\fbf$ by comparing predicted field $\zbf \in \C^M$ against the actual measurements $\ybf \in \C^M$.}
\label{Fig:NetworkRepresentation}
\end{figure}

%%%%%%%%%%%%%%%%%%%%%%%%%%%%%%%%%%%%%%%%%%%%%
%% Formulation with DNNs
%%%%%%%%%%%%%%%%%%%%%%%%%%%%%%%%%%%%%%%%%%%%%

\subsection{Network Interpretation}
\label{Sec:DNN}

We now discretize and combine equations~\eqref{Eq:ScatteredField} and~\eqref{Eq:IterativeBorn} into the following matrix-vector recursion
%\begin{subequations}
%\label{Eq:ForwardModel}
%\begin{align}
%&z_m \leftarrow \sum_{n = 1}^N H_{mn} u^K_n f_n \\
%&u^k_n \leftarrow u_n^0 + \sum_{i = 1}^N G_{n i } u_i^{k-1} f_i,
%%&z_m \leftarrow \sum_{n = 1}^N H_{mn} u^K_n f_n &\Leftrightarrow& \quad \zbf \leftarrow \Hbf (\ubf^K \odot \fbf), \\
%%&u^k_n \leftarrow u_n^0 + \sum_{i = 1}^N G_{n i } u_i^{k-1} f_i &\Leftrightarrow& \quad \ubf^k \leftarrow  \ubf^0 + \Gbf (\ubf^{k-1} \odot \fbf),
%\end{align}
%\end{subequations}
%where $k = 1,\dots, K$, $m = 1, \dots, M$, and $n = 1, \dots, N$. This is equivalently expressed in matrix-vector form as
\begin{subequations}
\label{Eq:MatVecForwardModel}
\begin{align}
&\zbf \leftarrow \Hbf (\ubf^K \odot \fbf), \\
&\ubf^k \leftarrow  \ubf^0 + \Gbf (\ubf^{k-1} \odot \fbf),
\end{align}
\end{subequations}
for $k = 1,\dots, K$. 
Here, the vector $\fbf \in \R^N$ is the discretization of the scattering potential $f$, $\zbf \in \C^M$ is the predicted scattered field $\uscat$ at sensor locations $\{\xbm_m\}_{m \in [1\dots M]}$, $\ubf^0 \in \C^N$ is the discretization of the input field $\uin$ inside $\Omega$, $\Hbf \in \C^{M \times N}$ is the discretization of the Green's function at sensor locations, $\Gbf \in \C^{N \times N}$ is the discretization of the Green's function inside $\Omega$, and $\odot$ denotes a component-wise multiplication between two vectors. For every $k \in [1\dots K]$, the vector $\ubf^k \in \C^N$ denotes the discretized version of the internal field after the $k$th scattering.

Figure~\ref{Fig:NetworkRepresentation} illustrates the representation
of~\eqref{Eq:MatVecForwardModel} as a feedforward neural
network~\cite{Bishop1995}, where the edge weights are represented in
$\Hbf$ and $\Gbf$ and the nonlinear nodes are described by the
scattering potential $\fbf$. Note that the linear edge weights
correspond to convolution operators and can thus be efficiently
implemented with FFTs. Accordingly, the total computational cost of
evaluating one forward pass through the network is $\Ocal(KN\log(N))$.

%%%%%%%%%%%%%%%%%%%%%%%%%%%%%%%%%%%%%%%%%%%%%
%% Inverse scattering
%%%%%%%%%%%%%%%%%%%%%%%%%%%%%%%%%%%%%%%%%%%%%

\subsection{Inverse Scattering}
\label{Sec:InverseScattering}

We formulate the inverse scattering as the following minimization problem
\begin{equation}
\label{Eq:OptimizationProblem}
\fbfhat = \argmin_{\fbf \in \Fcal} \{\Dcal(\fbf) + \tau \Rcal(\fbf)\},
\end{equation}
where $\Dcal$ and $\Rcal$ are the data-fidelity and regularization terms, respectively, and $\tau > 0$ is the regularization parameter.  The convex set $\Fcal \subseteq \R^N$ enforces physical constraints on the scattering potential such as its, for example, non-negativity. The data-fidelity term is given by
\begin{equation*}
\Dcal(\fbf) \defn \frac{1}{2}\|\ybf - \zbf(\fbf)\|_{\ell_2}^2,
\end{equation*}
where $\ybf \in \C^M$ contains measurements of the scattered field and $\zbf$ is the field predicted by the recursion~\eqref{Eq:MatVecForwardModel}. As a regularization term, we propose to use isotropic TV penalty~\cite{Rudin.etal1992}
\begin{equation*}
\Rcal(\fbf) \defn \sum_{n = 1}^N \|[\Dbf\fbf]_n\|_{\ell_2} = \sum_{n = 1}^N \sqrt{|[\Dbf_x \fbf]_n|^2+|[\Dbf_y\fbf]_n|^2},
\end{equation*}
where $\Dbf: \R^N \rightarrow \R^{N \times 2}$ is the discrete gradient operator with matrices $\Dbf_x$ and $\Dbf_y$ denoting the finite difference operations along $x$ and $y$ directions, respectively.

The optimization~\eqref{Eq:OptimizationProblem} can be performed iteratively using a proximal-gradient scheme or one of its accelerated variants~\cite{Bioucas-Dias.Figueiredo2007, Beck.Teboulle2009a, Kamilov2015a}. Specifically, the scattering potential can be updated with the following iteration
\begin{equation}
\label{Eq:ProxGradIteration}
\fbf^t \leftarrow \prox_{\gamma\tau \Rcal}\left(\fbf^{t-1} - \gamma \nabla \Dcal(\fbf^{t-1})\right),
\end{equation}
where $\gamma > 0$ is a step-size and
\begin{equation}
\prox_{\tau \Rcal} (\gbf) \defn \argmin_{\fbf \in \Fcal} \left\{\frac{1}{2}\|\fbf - \gbf\|_{\ell_2}^2 + \tau \Rcal(\fbf)\right\}
\end{equation} is the proximal operator, which corresponds to the TV regularized solution of the denoising problem. Note that, although, the proximal operator for isotropic TV does not admit a closed form, it can be efficiently computed~\cite{Beck.Teboulle2009a}. The gradient $\nabla \Dcal$ can be expressed as follows
\begin{equation}
\label{Eq:Gradient}
\nabla \Dcal(\fbf) = \real \left\{\left[\frac{\partial}{\partial \fbf} \zbf(\fbf)\right]^\Hrm (\zbf(\fbf)-\ybf)\right\},
\end{equation}
where $^\Hrm$ is the Hermitian transpose of the Jacobian
\begin{equation}
\label{Eq:Jacobian}
\frac{\partial}{\partial \fbf} \zbf(\fbf) \defn \left[\frac{\partial \zbf}{\partial f_1} \,\dots\, \frac{\partial \zbf}{\partial f_N}\right].
\end{equation}
Then, by differentiating equations in~\eqref{Eq:MatVecForwardModel} with respect to $\fbf$, and simplifying the resulting expressions, we obtain for any two vectors $\rbf \in \C^M$ and $\bbf \in \C^N$
\begin{align*}
&\left[\frac{\partial \zbf}{\partial \fbf}\right]^\Hrm \rbf = \left(\Hbf^\Hrm \rbf\right) \odot \ubfbar^K + \left[\frac{\partial \ubf^K}{\partial \fbf}\right]^\Hrm\left(\left(\Hbf^\Hrm \rbf\right) \odot \fbf\right) \\
&\left[\frac{\partial \ubf^k}{\partial \fbf}\right]^\Hrm \bbf = \left(\Gbf^\Hrm \bbf\right) \odot \ubfbar^{k-1} + \left[\frac{\partial \ubf^{k-1}}{\partial \fbf}\right]^\Hrm\left(\left(\Gbf^\Hrm \bbf\right) \odot \fbf\right),
\end{align*}
where $k = 1, \dots, K$, and       vector $\bar{\ubf}$ contains complex conjugated elements of $\ubf$. These relationships lead to the following error backpropagation algorithm
\begin{subequations}
\begin{align}
\gbf^k &\leftarrow \gbf^{k+1} + \left[\Gbf^\Hrm \vbf^{k+1}\right]\odot \ubfbar^k\\
\vbf^k &\leftarrow \left[\Gbf^\Hrm \vbf^{k+1}\right] \odot \fbf,
\end{align}
\end{subequations}
where $k = K-1, K-2, \dots, 0$, with the initialization ${\vbf^K = [\Hbf^\Hrm (\zbf-\ybf)]\odot \fbf}$ and ${\gbf^K = \left[\Hbf^\Hrm(\zbf-\ybf)\right] \odot \ubfbar^K}$. The final value of the gradient~\eqref{Eq:Gradient} is obtained by returning ${\nabla\Dcal(\fbf) = \real\{\gbf^0\}}$. 

The remarkable feature of our error backpropagation approach is that it allows to efficiently evaluate the gradient of the scattered field with respect to the scattering potential. Due to the convolutional structure of the matrices, its computational complexity is equivalent to running a forward pass, which is of order $\Ocal(KN\log(N))$. Equipped with this algorithm, the scattering potential can be optimized via iteration~\eqref{Eq:ProxGradIteration}. Note that the algorithm does not explicitly evaluate and store the Jacobian~\eqref{Eq:Jacobian} by instead computing its product with the residual $\rbf = (\zbf(\fbf) - \ybf)$, as indicated in~\eqref{Eq:Gradient}.

%%%%%%%%%%%%%%%%%%%%%%%%%%%%%%%%%%%%%%%%%%%%%
%% Section 4
%%%%%%%%%%%%%%%%%%%%%%%%%%%%%%%%%%%%%%%%%%%%%

\section{Experimental Validation}
\label{Sec:Experiments}

%%%%%%%%%%%%%%%%%%%%%%%%%%%%%%%%%%%%%%%%%%%%%
%% Results 1
%%%%%%%%%%%%%%%%%%%%%%%%%%%%%%%%%%%%%%%%%%%%%

\begin{figure}[t]
\centering\includegraphics[width=8.5cm]{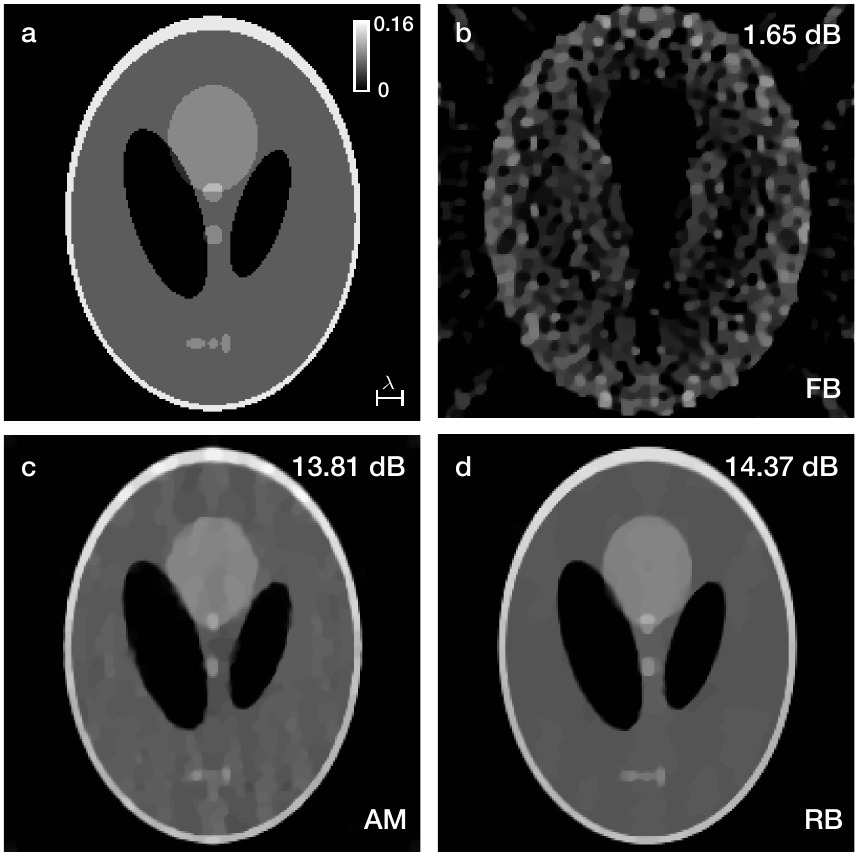}
\caption{Evaluation on \emph{Shepp-Logan} with $15$\% permittivity contrast at $\lambda = 7.49$ cm. (a) True contrast; (b) first-Born approximation; (c) alternating minimization; (d) proposed recursive-Born method with $K = 32$. Scale bar is equal to $\lambda$.}
\label{Fig:Results1}
\end{figure}

%%%%%%%%%%%%%%%%%%%%%%%%%%%%%%%%%%%%%%%%%%%%%
%% Results 2
%%%%%%%%%%%%%%%%%%%%%%%%%%%%%%%%%%%%%%%%%%%%%

\begin{figure}[t]
\centering\includegraphics[width=8.5cm]{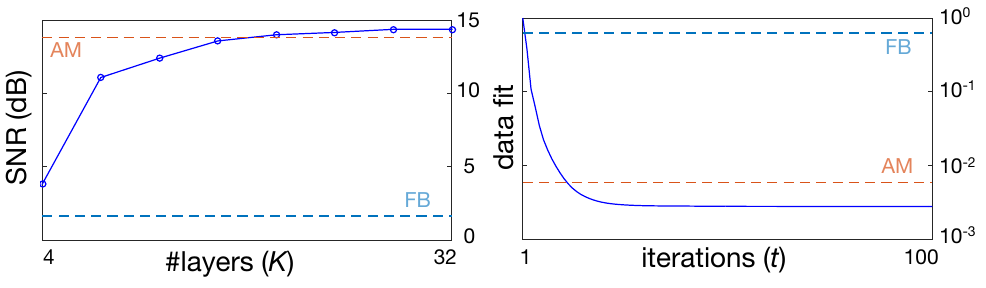}
\caption{Illustration of the reconstruction performance on \emph{Shepp-Logan} with $15$\% contrast. Left: SNR is plotted against the number of layers in the neural network. Right: normalized error between the true and predicted fields is plotted at each iteration for $K = 32$. Results obtained with FB and AM are marked with dashed lines.}
\label{Fig:Results2}
\end{figure}

To validate our \emph{Recursive Born (RB)} method, we report results for the tomographic experiment illustrated in Fig.~\ref{Fig:ScatteringScenario}, where the scattered wave measurements were obtained by running a high-fidelity Finite-Difference Time-Domain (FDTD)~\cite{Taflove.Hagness2005} simulator. The object is the \emph{Shepp-Logan phantom} of size $82$ cm $\times$ $112$ cm and the background medium is air with $\epsilon_b = 1$. The measurements are collected over $24$ transmissions on a circle of radius $R = 100$ cm with $15^{\text{o}}$ angle increments and, for each transmission, $360$ measurements around the object are recorded. The dimensions of the computational domain for reconstruction are set to $L_x = L_y = 120$ cm, with sampling steps $\delta x = \delta y = 0.6$ cm, which implies a measurement ratio of $M/N \approx 22\%$. We define the \emph{permittivity contrast} as $f_{\text{\tiny max}} \defn (\epsilon_{\text{\tiny max}}-\epsilon_b)/\epsilon_b$, where $\epsilon_{\text{\tiny max}} \defn \max_{\xbm \in \Omega} \{\epsilon(\xbm)\}$.

We compare results of our approach against two alternative methods. As the first reference method (denoted \emph{first Born (FB)}), we consider the TV-regularized solution of linearized model based on the first-Born approximation, which is known to be valid only for weakly scattering objects. In addition to the linearized approach, we consider an optimization scheme (denoted \emph{Alternating Minimization (AM)}) that alternates between updating the scattering potential for a fixed field and updating the field for a fixed scattering potential. This method is conceptually similar to the one proposed in~\cite{Wang.Chew1989}, but with TV replacing the smooth regularizer, due to the edge-preserving properties of the former. Hence, all three methods minimize the same TV-regularized least-squares error functional; however, each method relies on a distinct physical forward model. Additionally, all the methods were initialized with zero, iterated until convergence by measuring the change in two successive estimates, and their regularization parameters were set to the same value ${\tau = 10^{-9} \times \half \|\ybf\|_{\ell_2}^2}$.

Figure~\ref{Fig:Results1} compares the quality of the images obtained
by all three methods for the permittivity contrast of $15\%$ at
$\lambda = 7.49$ cm. Note that, due to the object's large size and
contrast, FB fails to characterize its structure. Alternatively, both
AM and RB succeed at recovering the object, with RB obtaining an image
of significantly superior quality. Figure~\ref{Fig:Results2}
illustrates the influence of the number of layers $K$ on the quality
of the reconstructed image (left), and the evolution of the relative
data-fit ${\|\ybf-\zbf\|^2_{\ell_2}/\|\ybf\|^2_{\ell_2}}$ for every
iteration with $K = 32$ (right). As can be appreciated from these
plots, the proposed method outperforms FB and AM, both in terms of
signal-to-noise ratio (SNR) and data-fit for networks with sufficient
number of layers. Additionally, the method converges relatively
fast---within first few tens of iterations.

%%%%%%%%%%%%%%%%%%%%%%%%%%%%%%%%%%%%%%%%%%%%%
%% Results 2
%%%%%%%%%%%%%%%%%%%%%%%%%%%%%%%%%%%%%%%%%%%%%

\begin{table}[t]
\def\arraystretch{1.15}%  1 is the default, change whatever you need
\small
\begin{center}
{\caption{Comparison of three methods in terms of SNR for $\lambda = 10$ cm. The best result for each contrast is highlighted.\label{Tab:Table}}}
\begin{tabular}{| l | c c c c  |}
\hline
& \multicolumn{4}{c |}{Permittivity Contrast $f_{\text{\tiny max}}$} \\
\cline{2-5}
& $5\%$ & $10\%$ & $15\%$ & $20\%$ \\
\hline
First Born & $9.20$ & $5.40$ & $2.98$ & $1.58$  \\
Alternating Minimization & $13.65$ & $13.61$ & $13.34$ & $12.98$  \\
Recursive Born & $\bm{14.86}$ & $\bm{14.07}$ & $\bm{13.71}$ & $\bm{13.65}$  \\
\hline
\end{tabular}
\end{center}
\end{table}

Table~\ref{Tab:Table} presents the results of quantitative evaluation of the methods for different values of the permittivity contrast at $\lambda = 10$ cm. As expected, the performance of all the methods degrades as the contrast value increases, which might be due to growing degree of nonlinearity, and hence, nonconvexity of the inverse scattering problem~\cite{Bucci.etal2001}. However, the solutions computed by the proposed RB approach are substantially better than the two alternatives, FB and AM, for all values of the contrast.

Finally, from a computational perspective, a single iteration of the
method requires a number of FFTs proportional to the number of
scattering layers. In our simulations, accurate results are obtained
using about 100 iterations with networks of approximately 20
layers. More concretely, our basic MATLAB implementation requires
about 1.3 seconds per iteration to process a transmission on a 4GHz
Intel Core i7 processor with 32 GBs of memory.

%%%%%%%%%%%%%%%%%%%%%%%%%%%%%%%%%%%%%%%%%%%%%
%% Section 5
%%%%%%%%%%%%%%%%%%%%%%%%%%%%%%%%%%%%%%%%%%%%%

\section{Conclusion}
\label{Sec:Conclusion}

The method developed in this paper reconstructs the distribution of the permittivity in an object from a set of measured scattered waves. In particular, the method accounts for multiple scattering of waves, in both transmission and reflection, and can thus be used when linearized models fail. The method is also computationally tractable due to its convolutional structure and can be further accelerated by parallelizing computations over multiple CPUs. We believe that the approach presented here opens rich perspectives for high-resolution tomographic imaging in a range of practical setups where multiple scattering is an issue.

%%%%%%%%%%%%%%%%%%%%%%%%%%%%%%%%%%%%%%%%%%%%%
%% References
%%%%%%%%%%%%%%%%%%%%%%%%%%%%%%%%%%%%%%%%%%%%%

\bibliographystyle{IEEEtran}

\begin{thebibliography}{10}
\providecommand{\url}[1]{#1}
\csname url@samestyle\endcsname
\providecommand{\newblock}{\relax}
\providecommand{\bibinfo}[2]{#2}
\providecommand{\BIBentrySTDinterwordspacing}{\spaceskip=0pt\relax}
\providecommand{\BIBentryALTinterwordstretchfactor}{4}
\providecommand{\BIBentryALTinterwordspacing}{\spaceskip=\fontdimen2\font plus
\BIBentryALTinterwordstretchfactor\fontdimen3\font minus
  \fontdimen4\font\relax}
\providecommand{\BIBforeignlanguage}[2]{{%
\expandafter\ifx\csname l@#1\endcsname\relax
\typeout{** WARNING: IEEEtran.bst: No hyphenation pattern has been}%
\typeout{** loaded for the language `#1'. Using the pattern for}%
\typeout{** the default language instead.}%
\else
\language=\csname l@#1\endcsname
\fi
#2}}
\providecommand{\BIBdecl}{\relax}
\BIBdecl

\bibitem{Kak.Slaney1988}
A.~C. Kak and M.~Slaney, \emph{Principles of Computerized Tomographic
  Imaging}.\hskip 1em plus 0.5em minus 0.4em\relax {IEEE}, 1988.

\bibitem{Wolf1969}
E.~Wolf, ``Three-dimensional structure determination of semi-transparent
  objects from holographic data,'' \emph{Opt. Commun.}, vol.~1, no.~4, pp.
  153--156, September/October 1969.

\bibitem{Devaney1981}
A.~J. Devaney, ``Inverse-scattering theory within the {R}ytov approximation,''
  \emph{Opt. Lett.}, vol.~6, no.~8, pp. 374--376, August 1981.

\bibitem{Bronstein.etal2002}
M.~M. Bronstein, A.~M. Bronstein, M.~Zibulevsky, and H.~Azhari,
  ``Reconstruction in diffraction ultrasound tomography using nonuniform
  {FFT},'' \emph{IEEE Trans. Med. Imag.}, vol.~21, no.~11, pp. 1395--1401,
  November 2002.

\bibitem{Sung.Dasari2011}
Y.~Sung and R.~R. Dasari, ``Deterministic regularization of three-dimensional
  optical diffraction tomography,'' \emph{J. Opt. Soc. Am. A}, vol.~28, no.~8,
  pp. 1554--1561, August 2011.

\bibitem{Kim.etal2014}
T.~Kim, R.~Zhou, M.~Mir, S.~Babacan, P.~Carney, L.~Goddard, and G.~Popescu,
  ``White-light diffraction tomography of unlabelled live cells,'' \emph{Nat.
  Photonics}, vol.~8, pp. 256--263, March 2014.

\bibitem{Simonetti2006}
F.~Simonetti, ``Multiple scattering: {T}he key to unravel the subwavelength
  world from the far-field pattern of scattered wave,'' \emph{Phys. Rev. E:
  Stat., Nonlinear, Soft Matter Phys.}, vol.~73, no.~3, p. 036619, March 2006.

\bibitem{Simonetti.etal2008}
F.~Simonetti, M.~Fleming, and E.~A. Marengo, ``Illustration of the role of
  multiple scattering in subwavelength imaging from far-field measurements,''
  \emph{J. Opt. Soc. Am. A}, vol.~25, no.~2, pp. 292--303, February 2008.

\bibitem{Maire.etal2009}
G.~Maire, F.~Drsek, J.~Girard, H.~Giovaninni, A.~Talneau, D.~Konan,
  K.~Belkebir, P.~C. Chaumet, and A.~Sentenac, ``Experimental demonstration of
  quantitative imaging beyond {A}bbe's limit with optical diffraction
  tomography,'' \emph{Phys. Rev. Lett.}, vol. 102, p. 213905, May 2009.

\bibitem{Kamilov.etal2015}
U.~S. Kamilov, I.~N. Papadopoulos, M.~H. Shoreh, A.~Goy, C.~Vonesch, M.~Unser,
  and D.~Psaltis, ``Learning approach to optical tomography,'' \emph{Optica},
  vol.~2, no.~6, pp. 517--522, June 2015.

\bibitem{vandenBerg2002}
P.~M. {van den Berg}, \emph{Scattering}.\hskip 1em plus 0.5em minus 0.4em\relax
  Academic Press, 2002, ch. Nonlinear Scalar Inverse Scattering: Algorithms and
  Applications, pp. 142--161.

\bibitem{Tijhuis1989}
A.~G. Tijhuis, ``{B}orn-type reconstruction of material parameters of an
  inhomogeneous, lossy dielectric slab from reflected-field data,'' \emph{Wave
  Motion}, vol.~11, no.~2, pp. 151--173, May 1989.

\bibitem{Wang.Chew1989}
Y.~M. Wang and W.~Chew, ``An iterative solution of the two-dimensional
  electromagnetic inverse scattering problem,'' \emph{Int. J. Imag. Syst
  Tech.}, vol.~1, pp. 100--108, 1989.

\bibitem{Chew.Wang1990}
W.~C. Chew and Y.~M. Wang, ``Reconstruction of two-dimensional permittivity
  distribution using the distorted {B}orn iterative method,'' \emph{IEEE Trans.
  Med. Imag.}, vol.~9, no.~2, pp. 218--225, June 1990.

\bibitem{Kleinman.vandenBerg1992}
R.~E. Kleinman and P.~M. {van den Berg}, ``A modified gradient method for
  two-dimensional problems in tomography,'' \emph{J. Comput. Appl. Math.},
  vol.~42, no.~1, pp. 17--35, 1992.

\bibitem{Kleinman.vandenBerg1993}
------, ``An extended range-modified gradient technique for profile
  inversion,'' \emph{Radio Sci.}, vol.~28, no.~5, pp. 877--884,
  September-October 1993.

\bibitem{Belkebir.Sentenac2003}
K.~Belkebir and A.~Sentenac, ``High-resolution optical diffraction
  microscopy,'' \emph{J. Opt. Soc. Am. A}, vol.~20, no.~7, pp. 1223--1229, July
  2003.

\bibitem{vandenBerg.Kleinman1997}
P.~M. {van den Berg} and R.~E. Kleinman, ``A contrast source inversion
  method,'' \emph{Inv. Probl.}, vol.~13, no.~6, pp. 1607--1620, December 1997.

\bibitem{Tian.Waller2015}
L.~Tian and L.~Waller, ``{3D} intensity and phase imaging from light field
  measurements in an {LED} array microscope,'' \emph{Optica}, vol.~2, pp.
  104--111, 2015.

\bibitem{Waller.Tian2015}
L.~Waller and L.~Tian, ``Machine learning for {3D} microscopy,'' \emph{Nature},
  vol. 523, no. 7561, pp. 416--417, July 2015.

\bibitem{Kamilov.etal2016}
U.~S. Kamilov, I.~N. Papadopoulos, M.~H. Shoreh, A.~Goy, C.~Vonesch, M.~Unser,
  and D.~Psaltis, ``Optical tomographic image reconstruction based on beam
  propagation and sparse regularization,'' \emph{IEEE Trans. Comp. Imag.},
  vol.~2, no.~1, pp. 59--70, March 2016.

\bibitem{Born.Wolf2003}
M.~Born and E.~Wolf, \emph{Principles of Optics}, 7th~ed.\hskip 1em plus 0.5em
  minus 0.4em\relax Cambridge Univ. Press, 2003, ch. Scattering from
  inhomogeneous media, pp. 695--734.

\bibitem{Arfken.Weber2005a}
G.~B. Arfken and H.~J. Weber, \emph{Mathematical Methods for Physicists},
  6th~ed.\hskip 1em plus 0.5em minus 0.4em\relax Elsevier, 2005, ch. Integral
  equations, pp. 1005--1036.

\bibitem{Bishop1995}
C.~M. Bishop, \emph{Neural Networks for Pattern Recognition}.\hskip 1em plus
  0.5em minus 0.4em\relax Oxford, 1995.

\bibitem{Rudin.etal1992}
L.~I. Rudin, S.~Osher, and E.~Fatemi, ``Nonlinear total variation based noise
  removal algorithms,'' \emph{Physica D}, vol.~60, no. 1--4, pp. 259--268,
  November 1992.

\bibitem{Bioucas-Dias.Figueiredo2007}
J.~M. Bioucas-Dias and M.~A.~T. Figueiredo, ``A new {T}w{IST}: {T}wo-step
  iterative shrinkage/thresholding algorithms for image restoration,''
  \emph{IEEE Trans. Image Process.}, vol.~16, no.~12, pp. 2992--3004, December
  2007.

\bibitem{Beck.Teboulle2009a}
A.~Beck and M.~Teboulle, ``Fast gradient-based algorithm for constrained total
  variation image denoising and deblurring problems,'' \emph{IEEE Trans. Image
  Process.}, vol.~18, no.~11, pp. 2419--2434, November 2009.

\bibitem{Kamilov2015a}
U.~S. Kamilov, ``Parallel proximal methods for total variation minimization,''
  in \emph{{IEEE} Int. Conf. Acoustics, Speech and Signal Process. ({ICASSP}
  2016)}, Shanghai, China, March 19-25, 2015, arXiv:1510.00466 [cs.IT].

\bibitem{Taflove.Hagness2005}
A.~Taflove and S.~C. Hagness, \emph{Computational {E}lectrodynamics: The
  {F}inite-{D}ifference {T}ime-{D}omain Method}, 3rd~ed.\hskip 1em plus 0.5em
  minus 0.4em\relax Artech House Publishers, 2005.

\bibitem{Bucci.etal2001}
O.~M. Bucci, L.~Cardace, L.~Crocco, and T.~Isernia, ``Degree of nonlinearity
  and a new solution precudure in scalar two-dimensional inverse scattering
  problems,'' \emph{J. Opt. Soc. Am. A}, vol.~18, no.~8, pp. 1832--1843, August
  2001.

\end{thebibliography}


\end{document}